\documentclass[final]{cvpr}

\usepackage{times}
\usepackage{epsfig}
\usepackage{graphicx}
\usepackage{amsmath}
\usepackage{amssymb}

\usepackage{booktabs}
\usepackage{multirow}
\usepackage[ruled]{algorithm2e}

\usepackage[pagebackref=true,breaklinks=true,colorlinks,bookmarks=false]{hyperref}

\def\cvprPaperID{xxxx} 
\def\confYear{CVPR 2021}

\begin{document}

\title{Privileged Knowledge Distillation for Online Action Detection}
                  
\author{Peisen Zhao\\
Shanghai Jiao Tong University\\
{\tt\small pszhao@sjtu.edu.cn}
\and
Jiajie Wang\\
Alibaba Group\\
{\tt\small wjj171584@alibaba-inc.com}
\and
Lingxi Xie\\
Huawei Inc.\\
{\tt\small 198808xc@gmail.com}
\and
Ya Zhang\\
Shanghai Jiao Tong University\\
{\tt\small ya\_zhang@sjtu.edu.cn}
\and
Yanfeng Wang\\
Shanghai Jiao Tong University\\
{\tt\small wangyanfeng@sjtu.edu.cn}
\and
Qi Tian\\
Huawei Cloud \& AI\\
{\tt\small tian.qi1@huawei.com}
}

\maketitle

\begin{abstract}
   
Online Action Detection (OAD) in videos is proposed as a per-frame labeling task to address the real-time prediction tasks that can only obtain the previous and current video frames. This paper presents a novel learning-with-privileged based framework for online action detection where the future frames only observable at the training stages are considered as a form of privileged information. Knowledge distillation is employed to transfer the privileged information from the offline teacher to the online student. We note that this setting is different from conventional KD because the difference between the teacher and student models mostly lies in input data rather than the network architecture. We propose Privileged Knowledge Distillation (PKD) which (i) schedules a curriculum learning procedure and (ii) inserts auxiliary nodes to the student model, both for shrinking the information gap and improving learning performance. Compared to other OAD methods that explicitly predict future frames, our approach avoids learning unpredictable unnecessary yet inconsistent visual contents and achieves state-of-the-art accuracy on two popular OAD benchmarks, TVSeries and THUMOS14.
\end{abstract}

\section{Introduction}

Action detection in videos has been widely studied under offline settings \cite{chao2018rethinking, lin2018bsn, lin2019bmn, liu2019multi, gao2020accurate, zhao2020bottom}, where the entire video is observed when locating the actions. In some real-time situations such as autonomous driving and online surveillance where the video frames come in streams, real-time predictions based on the available data are required. To address this problem, Online Action Detection (OAD) is proposed as a per-frame labeling task with existing observations. 
\begin{figure}[t]
\begin{center}
  \includegraphics[width=1.0\linewidth]{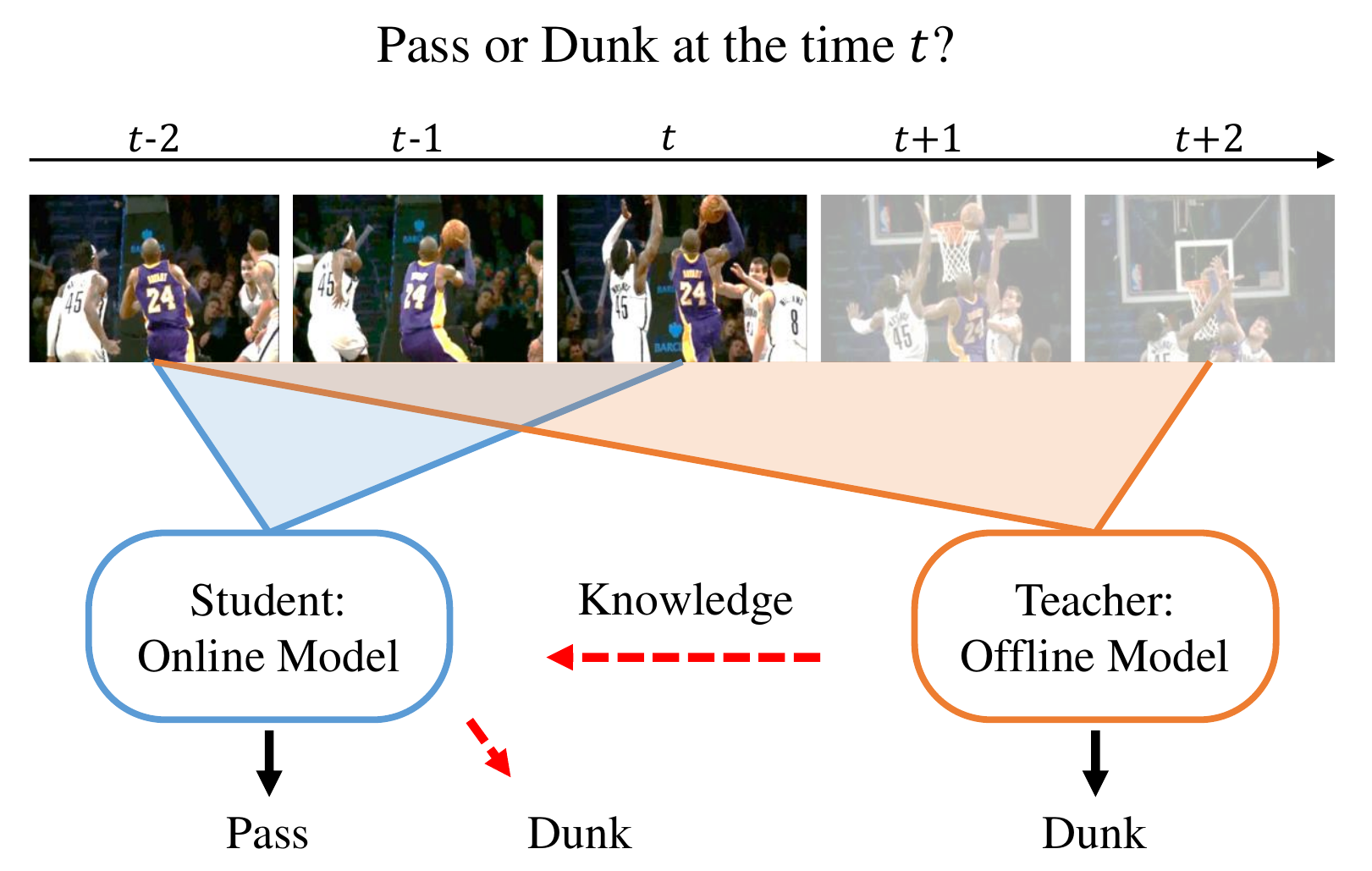}
\end{center}
\caption{``Pass or Dunk?'' When watching the basketball game at time \textit{t}, we prefer that Kobe will dunk since we may have seen him completing dunk before. This experience helps us make decisions. Inspired by this, it is beneficial to teach the OAD model by knowledge distillation with a teacher who can see full video data. 
}
\label{fig:causalconv}
\end{figure}

Early OAD methods \cite{geest2016online, shou2018online} only use the available video frames (\ie, previous and current frames) to train the online models.
Considering that future observations are actually available at the training stage,  F\textsuperscript{2}G \cite{yoon2020a} and TRN \cite{xu2019trn} propose to first learn to anticipate either future video frames or future frame features; then utilize these anticipations together with the existing observations to better classify actions. This anticipation\&classification-based decision mechanism has been shown to improve the performance of OAD. However, the anticipation may have a different target than the classification. Predicting the detailed visual appearance or general features extracted with a pre-trained model may not correlate well with the action semantics. 
Furthermore, the anticipation\&classification framework inevitably suffers from the efficiency issue, \ie, making explicit predictions especially predicting future video frames needs considerable computations and is not suitable for online tasks. 





We deem the additional future frames available at the training stage as \emph{privileged information}\cite{vapnik2009new, vapnik2015learning} and reformulate the problem into online action detection with privileged information. To leverage the privileged information but not explicitly anticipate it as the anticipation\&classification methods do, we introduce a Knowledge Distillation (KD) based framework, named Privileged Knowledge Distillation (PKD), 
which consists of an offline teacher and an online student, and anticipation is replaced with knowledge distillation from offline teacher models to online student models. The KD process can be seen as an implicit way of learning to anticipate high-level semantic features from offline models. With this KD framework, both the teacher and student have the same learning target of classifying current actions. To the best of our knowledge, this is the first study that introduces KD for the OAD task.

The KD for OAD is different from conventional KD because the difference between the teacher and student models mostly lies in the available input data rather than the network architecture. Given the different inputs, it is not feasible for the student to learn the same representation as teachers. Additional auxiliary nodes are thus inserted to each layer of the student network so that only part of the student features are regularized by their teachers.
Moreover, the information gap between the teacher and the student may not be easily captured through the knowledge distillation, especially at the early stage of an action.
To facilitate the information propagation from the teacher to the student, a curriculum learning procedure is further introduced. Instead of directly letting the student learn from the teacher, we formulate several intermediate teacher models that can see increasing lengths of future frames, representing teachers with different levels of knowledge. The student then learns from these intermediate teachers in the order of from easy to hard. In this way, the privileged knowledge from offline models is gradually distilled to the online student model.

We validate the effectiveness of the proposed PKD on two popular benchmarks TVSeries and THUMOS14. The experimental results show that our proposed auxiliary nodes and curriculum learning procedure indeed improves the performance of the OAD model and achieves the state-of-the-art performance of $86.44\%$ mcAP on TVSeries and $64.45\%$ mAP on THUMOS14. Further analysis shows that PKD performs better when detecting action at some early stages.

\section{Related Work}

\noindent \textbf{Offline Action Detection.} Offline action detection aims to detect the start and end of each action instance with fully-observed videos. Existing methods, such as TAL-Net \cite{chao2018rethinking}, BSN  \cite{lin2018bsn}, BMN \cite{lin2019bmn}, MGG \cite{liu2019multi}, and the most recent works \cite{gao2020accurate, zhao2020bottom} are inspired by the two-stage object detection framework \cite{ren2017faster}, that decomposed this task into the action proposal and action classification stages. Besides, AFO-TAD \cite{tang2019afotad} and A2Net \cite{yang2020revisiting} proposed an anchor-free method that directly regress the action boundaries without pre-defined action proposals. PGCN \cite{zeng2019graph} and G-TAD \cite{xu2020gtad} introduced GCN \cite{kipf2017semi} to model the relationship between action proposals and temporal nodes. All these offline methods need to observe the entire video.

\vspace{4pt}
\noindent \textbf{Online Action Detection.} Different from the offline settings, online action detection can only observe the current and previous video frames. Geest \emph{et al.} \cite{geest2016online} first defined the online action detection task and introduced a dataset, TVSeries. Gao \emph{et al.} \cite{gao2017red} proposed a reinforced encoder-decoder network to anticipate future actions that can be used for online action detection. TRN \cite{xu2019trn} also predicted future information and proposed a TRN cell to use these predicted features as well as the current and past features to detect actions. Besides, Shou \emph{et al.} \cite{shou2018online} proposed a new metric to evaluate the quality of action start points and named this task as Online Detection of Action Start (ODAS). StartNet \cite{gao2019startnet} followed the new metric and addressed ODAS by optimizing long-term localization rewards using policy gradient methods. Recently, \cite{eun2020learning} changed the state update rules in LSTM and proposed an Information Discrimination Unit (IDU) to model the relationship between the current frame and the past frames. Then the IDU could decide whether to accumulate input information based on its relevance to the current feature. Most of the online action detection methods are designed based on RNN models that are more suitable for modeling sequence data.

\vspace{4pt}
\noindent \textbf{Knowledge Distillation.} Ordinarily, knowledge distillation is an effective technique that can transfer the knowledge from teachers, \emph{i.e.}, large complex models, to students, \emph{i.e.}, small simple models. After Hinton \emph{et al.} \cite{hinton2015distilling} addressed knowledge distillation by training the student with the soft target provided by the teacher, knowledge distillation has been widely adopted in a variety of learning tasks. \cite{ba2014do, ahn2019variational, cho2019on, hegde2020variational, yang2019snapshot, liu2019knowledge} let the student imitate the output logits of the teacher; \cite{romero2014fitnets, tung2019similarity, shen2019meal, aguilar2019knowledge, gao2020residual, xu2020kernel} were focused on minimizing the representations between intermediate layers of the student and teacher.
Besides, some studies introduced the idea of progressive training. TAKD \cite{mirzadeh2020improved} and DGKD \cite{son2020densely} introduced medium size models, called teacher assistant, to bridge the architecture gap between the teacher and student models. 


\begin{figure*}[t]
\begin{center}
  \includegraphics[width=1\linewidth]{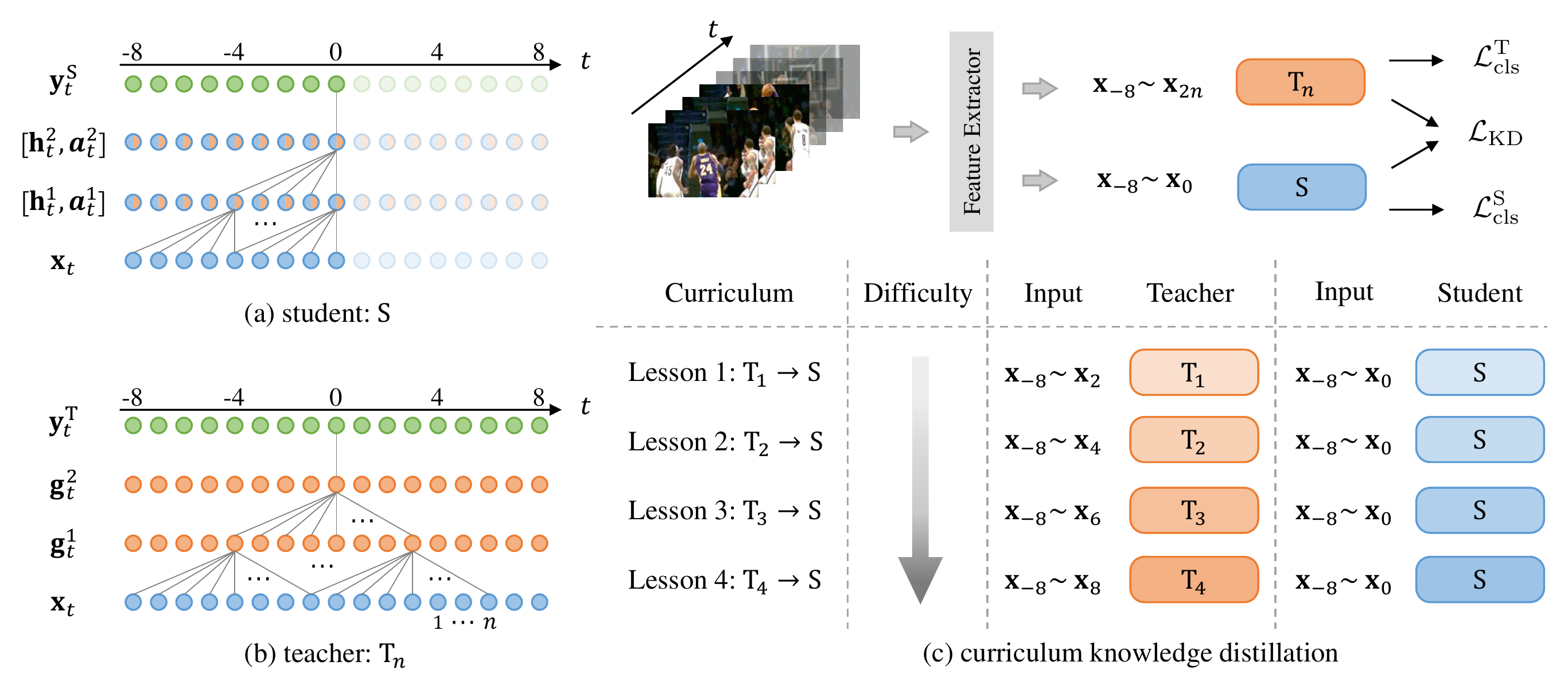}
\end{center}
\caption{(a) and (b) show the model structures of the student $\mathtt{S}$ and teacher $\mathtt{T}_{4}$. Both of them use the 1D temporal convolution layers. $\mathbf{x}_{t}$ is the extracted video features. $\mathbf{h}_{t}^{1}$ and $\mathbf{h}_{t}^{2}$ denote original student features; $\mathbf{a}_{t}^{1}$ and $\mathbf{a}_{t}^{2}$ denote auxiliary nodes; $\mathbf{g}_{t}^{1}$ and $\mathbf{g}_{t}^{2}$ denote teacher features. $\mathbf{y}_{t}^\mathrm{S}$ and $\mathbf{y}_{t}^\mathrm{T}$ are output logits. (c) shows the curriculum learning procedure that the student model $\mathtt{S}$ are trained from $\mathtt{T}_{1}$ to $\mathtt{T}_{4}$. $\mathtt{T}_{1}$, $\mathtt{T}_{2}$, $\mathtt{T}_{3}$ are also teacher models that can see different length of future data between $\mathtt{S}$ and $\mathtt{T}_{4}$.}
\label{fig:pkd}
\end{figure*}

\section{Preliminaries}

\noindent \textbf{Models and Notations.} Most OAD methods \cite{gao2019startnet, xu2019trn, eun2020learning} adopt the LSTM-based network to model the input sequence data. However, the LSTM structure is less efficient than the CNN based network. Thus, we adopt the causal convolution \cite{oord2016wavenet} to build our OAD model. As illustrated in Figure~\ref{fig:pkd} (a), a two-layer causal convolution model, named $\mathtt{S}$, can only observe the previous and current features, which satisfies the online settings. Besides, we also introduce an offline model to teach the student model $\mathtt{S}$ in Figure~\ref{fig:pkd} (b). This model does not only keep the same receptive fields in previous observations with the online model but also obtains the future observations. In terms of the extra increasing numbers of connections from the future nodes in convolution kernels, the offline teacher models are named $\mathtt{T}_{1}$, $\mathtt{T}_{2}$, $\mathtt{T}_{3}$, and $\mathtt{T}_{4}$.

\noindent \textbf{Problem Definition.} Given a streaming video, the goal of OAD is to recognize the current action without access to future frames.
We denote $\mathbf{x}_{t}\in\mathbb{R}^{C}$ as the input features to represent the streaming video and $\mathbf{y}_{t}\in\mathbb{R}^{M+1}$ as the classification logits. At the time step $t=0$ with the two-layer models shown in Figure~\ref{fig:pkd}, online model $\mathtt{S}$ and offline model $\mathtt{T}_{4}$ are formulated as:
$$
    \mathbf{y}_{0}^\mathrm{S}=\mathtt{S}(\{\mathbf{x}_{t}\}_{t=-8}^{0}), \quad
    \mathbf{y}_{0}^\mathrm{T}=\mathtt{T}_{4}(\{\mathbf{x}_{t}\}_{t=-8}^{8}).
$$
where $\mathbf{y}_{0}^\mathrm{S}\in\mathbb{R}^{M+1}$ and $\mathbf{y}_{0}^\mathrm{T}\in\mathbb{R}^{M+1}$ are the output logits at the time step $t=0$; $M$ denotes the number of action classes and $+1$ means the background class. Both the student and teacher models are supervised by classification labels $\mathbf{y}_{t}^\mathrm{gt}$:
\begin{equation}\label{eq:loss1}
    \mathcal{L}_\mathrm{cls}=\frac{1}{N}\sum_{t\in\mathcal{N}}\mathcal{H}_\mathrm{CE}(\mathbf{y}_{t}^\mathrm{gt},{\mathbf{y}_{t}}),
\end{equation}
where $N$ is the number of time steps; $\mathcal{H}_\mathrm{CE}$ represents the cross-entropy. The goal of KD for OAD is to distill the privileged knowledge from $\mathtt{T}_{4}$ to $\mathtt{S}$.

\section{Privileged KD for OAD}

Knowledge distillation is an effective technique that can transfer knowledge from the teacher to the student. It is natural to introduce KD to the OAD task which can distill the knowledge from the offline to online models. However, different from the conventional KD that both the teacher and student take the same input data, the KD in OAD settings takes the different inputs for teacher and student models. Therefore, the main difference between the teacher and student comes from the input information gap rather than model structures. This information gap makes it a challenge to let the student learn from the teacher with existing KD methods. Thus, we propose a novel Privileged Knowledge Distillation (PKD) method to cushion the information gap between them by (i) scheduling a \emph{curriculum learning} process for KD and (ii) inserting \emph{auxiliary nodes} for the student model. These designs are illustrated in Sections 4.1 and 4.2, and we summarize the loss function in Section 4.3.

\subsection{Curriculum Knowledge Distillation}

Curriculum Learning \cite{bengio2009curriculum} is proposed by Bengio \emph{et al.} that teaches a model starting from easy samples to hard samples. Inspired by this training strategy, we introduce a curriculum learning procedure to PKD that can cushion the information gap between the teacher and student. In this section, we first introduce the Vanilla KD and then describe the curriculum learning procedure.

\vspace{4pt}

\noindent \textbf{Vanilla KD.} 
The output logits from a classification model can be regarded as a soft label that contains the knowledge of how similar between different classes. This similarity knowledge is helpful for classification tasks, which is proved by previous KD methods \cite{hinton2015distilling,ahn2019variational,hegde2020variational}. With observing the future data, logits from the offline model contain a more complete understanding of this similarity knowledge than logits from the online model. So we train the online model to imitate the output logits of the offline model by minimizing the KL divergence between them at each time step $t$. Therefore, the loss function of the vanilla knowledge distillation is formulated as:  
\begin{equation}\label{eq:loss2}
    \mathcal{L}_{\mathrm{L}}=\frac{1}{N}\sum_{t}\tau^{2} \mathcal{L}_\mathrm{KL}(\sigma(\mathbf{y}_{t}^\mathrm{S}/\tau), \sigma(\mathbf{y}_{t}^\mathrm{T}/\tau)),
\end{equation}
where $N$ is the number of time steps; $\tau$ is the temperature parameter \cite{hinton2015distilling} to control the softening of $\mathbf{y}_\mathrm{T}$; $\sigma$ is the softmax function; $\mathcal{L}_\mathrm{KL}(\cdot,\cdot)$ measures the KL divergence.

\vspace{4pt}

\noindent \textbf{Curriculum Learning Procedure.} 
The information gap between $\mathtt{S}$ and $\mathtt{T}_{4}$ is large since half of the input data is not available. We find it is difficult to directly train the student model $\mathtt{S}$ to learn from $\mathtt{T}_{4}$. Therefore, to better reduce the information gap between the teacher and student, we schedule a curriculum learning procedure to relieve this problem. As illustrated in Figure~\ref{fig:pkd} (c), we first introduce some intermediate teacher models $\mathtt{T}_{1}$, $\mathtt{T}_{2}$, and $\mathtt{T}_{3}$ which can see the increasing length of future frames between $\mathtt{S}$ and $\mathtt{T}_{4}$; then let the student model $\mathtt{S}$ learn from the teacher models from $\mathtt{T}_{1}$ to $\mathtt{T}_{4}$. This learning process is notated as $\mathtt{S}\to\mathtt{T}_{1}\to\mathtt{T}_{2}\to\mathtt{T}_{3}\to\mathtt{T}_{4}$ and the detailed training procedure is shown in Algorithm~\ref{alg:bwpkd}. In addition, TAKD \cite{mirzadeh2020improved} also uses intermediate teacher models to help KD. But the distillation path is opposite from ours, so it will lose the input information in the final distillation stage when training the student model. The detailed comparisons are shown in the experimental results.

\begin{algorithm}[t]
\caption{Curriculum Learning Procedure.}
\label{alg:bwpkd}
\LinesNumbered
\KwIn {extracted video features $\mathbf{x}_{t}$, target labels $\mathbf{y}_{t}^\mathrm{gt}$}
\KwOut {distilled student model $\mathtt{S}$}
Train the student $\mathtt{S}$ with Eq.~\eqref{eq:loss1}

Train the teachers $\mathtt{T}_{1}$, $\mathtt{T}_{2}$, $\mathtt{T}_{3}$, and $\mathtt{T}_{4}$ with Eq.~\eqref{eq:loss1}

\For{$i\gets1$ \KwTo $4$}{
    load the parameters of the models $\mathtt{S}$, $\mathtt{T}_{i}$

    \SetKwRepeat{Do}{do}{while}
        \Do{$\mathtt{S}$ is not converged}{ 
        inference the model $\mathtt{S}$ and $\mathtt{T}_{i}$ with $\mathbf{x}_{t}$
        
        compute the gradients with Eq.~\eqref{eq:loss5}
        
        update the parameters of $\mathtt{S}$ and $\mathtt{T}_{i}$
        }
    save the parameters of the model $\mathtt{S}$
    
}

\end{algorithm}

\subsection{Learning with Auxiliary Nodes}

Conventional KD methods usually force the student model to have the same representations as the teacher model does. It is appropriate to add this constrain since they have the same input data. However, with different input data, it is difficult to let the student model $\mathtt{S}$ to learn the same representations as the teacher does. Therefore, directly minimizing the distance between their feature representations is not appropriate in this setting. So we introduce a new KD method that inserts auxiliary nodes for the student to relieve this problem. As illustrated in Figure~\ref{fig:pkd} (a), each two-color node in the feature layer represents a combined feature representation that concatenates the original student features $\mathbf{h}_{t}^{l}\in\mathbb{R}^{C^{l}}$ with the auxiliary nodes $\mathbf{a}_{t}^{l}\in\mathbb{R}^{C^{l}}$. $l$ stands for the feature layer and the $C^{l}$ represents the feature dimension. At the training stage, we let these auxiliary nodes learn from the teacher representations $\mathbf{g}_{t}^{l}\in\mathbb{R}^{C^{l}}$ which are shown in Figure~\ref{fig:pkd} (b) by orange nodes. In this case, the auxiliary nodes can serve as the additional information that will not conflict with original feature representations of the student model and can also provide privileged knowledge. 

\noindent \textbf{Auxiliary Nodes for Final Feature Layer.} 
At the time step $t$ in Figure~\ref{fig:pkd} (a), auxiliary nodes $\mathbf{a}_{t}^{2}$ in the final feature layer are used to classify actions at that moment. So these auxiliary nodes of the final feature layer should be supervised by the corresponding teacher representations $\mathbf{g}_{t}^{2}$. Thus, the loss function is formulated as:
\begin{equation}\label{eq:loss3}
    \mathcal{L}_{\mathrm{AN}}=\frac{1}{N}\sum_{t}||\mathbf{a}_{t}^{2}-\mathbf{g}_{t}^{2}||^{2},
\end{equation}
where $N$ is the number of time steps and $||\cdot||^{2}$ measures the mean square error.

\noindent \textbf{Auxiliary Nodes for Intermediate Feature Layer.} 
Different from the conventional KD with features, directly applying Eq.~\eqref{eq:loss3} for intermediate feature layers makes an unmatched-receptive-field problem. It is because that the teacher and student models use the different convolution types. Specifically, at the time step $t=0$ in layer $2$, the student feature $\mathbf{d}_{0}^{2}$ has the receptive fields of $[-8,0]$, while the teacher feature $\mathbf{g}_{0}^{2}$ has the receptive fields of $[-8,-8]$ at the same position. However, at layer $1$, the corresponding features $\{\mathbf{d}_{t}^{1}\}_{t=-4}^{0}$ in $\mathtt{S}$ have the same receptive fields of $[-8,0]$ but $\{\mathbf{g}_{t}^{1}\}_{t=-4}^{0}$ in $\mathtt{T}_{4}$ have the receptive fields of $[-8,+4]$ in the same position. To solve this unmatched-receptive-field problem, we let the student features $\mathbf{d}_{t}^{1}$ at layer 1 to learn from the $\mathbf{g}_{t}^{1}$ with seeing more future nodes to align the receptive fields to $[-8,+8]$. So the final loss function of the KD with AN is formulated as:
\begin{equation}\label{eq:loss4}
    \mathcal{L}_{\mathrm{AN}}=\frac{1}{N}\sum_{t}||\mathbf{d}_{t}^{2}-\mathbf{g}_{t}^{2}||^{2}+||\mathbf{d}_{t}^{1}-\frac{1}{p+1}\sum_{t}^{t+p}\mathbf{g}_{t}^{1}||^{2},
\end{equation}
where $N$ is the number of time steps and $p$ is the number of future nodes that the model should be seen to align the receptive fields. $p$ is set to be 4 for KD from T4 to $\mathtt{S}$.




        
        
    


\subsection{Loss Function}

Instead of using the fixed teacher models, we jointly optimize the student and teacher models to obtain a better performance which has been confirmed by \cite{zhang2018deep}. Therefore, using the Eq.~\eqref{eq:loss1}, Eq.~\eqref{eq:loss2}, and Eq.~\eqref{eq:loss4}, the overall loss function for training the OAD model with KD is formulated as:
\begin{equation}\label{eq:loss5}
    \mathcal{L}=\mathcal{L}_\mathrm{cls}+\mathcal{L}_\mathrm{KD}=\mathcal{L}_\mathrm{cls}^\mathrm{S}+\mathcal{L}_\mathrm{cls}^\mathrm{T}+\lambda\mathcal{L}_\mathrm{L}+\alpha\mathcal{L}_\mathrm{AN},
\end{equation}
where $\mathcal{L}_\mathrm{cls}^\mathrm{S}$ and $\mathcal{L}_\mathrm{cls}^\mathrm{T}$ are the classification losses for the student and teacher models. $\mathcal{L}_\mathrm{L}$ denotes the vanilla KD loss while $\mathcal{L}_\mathrm{AN}$ is the KD loss for auxiliary nodes. $\lambda$ and $\alpha$ are the trade-off parameters.

\section{Experiments}

\subsection{Datasets and Evaluation Metric}

\noindent \textbf{TVSeries.} The TVSeries dataset \cite{geest2016online} consists of $16$ hours of $27$ untrimmed videos with $30$ action classes, \emph{e.g.}, ``stand up'', ``open door'', and ``drive car''. The dataset is collected from the six recent TV series on DVD. So it is challenging with diverse actions, multiple actors, heavy occlusions, and a large proportion of background video frames. 

\noindent \textbf{THUMOS14.} The THUMOS14 dataset includes $413$ untrimmed videos with $20$ action classes. Following the public data split, $200$ of them are used for training and $213$ are used for testing. Each video contains more than an average of $15$ action instances and $70\%$ backgrounds.

\noindent \textbf{Evaluation Metric.}
To evaluate the performance of the online action detection, we follow the previous studies~\cite{xu2019trn,eun2020learning} and use the conventional metric mean average precision (mAP), which first calculate the average precision (AP) of all frames for each action class; then average these APs over all classes. The mAP is used for the THUMOS14 dataset. As for TVSeries, \cite{geest2016online} proposed a mean calibrated average precision (mcAP) to better evaluate the performance of the online action detection, which balance the effectiveness of the foreground-background proportion. The calibrated average precision is formulated as:
\begin{equation}\label{eq:metric}
    \mathrm{cAP} = \frac{\sum_{k}\mathrm{cPrec}(k)\times\mathrm{I}(k)}{P},
\end{equation}
where the calibrated precision $\mathrm{cPrec}=\frac{\mathrm{TP}}{\mathrm{TP}+\mathrm{FP}/\omega}$; $\mathrm{I}(k)$ is equal to 1 if frame k is a True Positive (TP); $P$ is the number of TP. The coefficient $\omega$ is the ratio between the number of background and action frames. Thus, the mcAP is calculated by averaging cAPs over all classes.

\subsection{Implementation Details}

We use the Pytorch framework \cite{paszke2017automatic} to implement our online action detection models as well as the training process. All the proposed models $\mathtt{T}_{1}$-$\mathtt{T}_{4}$ and $\mathtt{S}$ are trained with the Adam optimizer. All the experiment results are the averaged values of $3$ times.

\noindent \textbf{Parameter settings.}
At each training process including training the single model (\emph{e.g.}, $\mathtt{T}_{4}$ or $\mathtt{S}$) or training with KD loss (\emph{e.g.}, distilling $\mathtt{T}_{4}$ to $\mathtt{S}$), the number of epochs is set to be $40$. The initial learning rate is set to be $5\times10^{-4}$ and decreased by a factor of $0.1$ at the epoch 30. The hyper-parameter of temperature $\tau$ is set to be $5$. The trade-off parameter $\lambda$ and $\alpha$ are set to be $0.4$ and $0.01$ that can balance the KD loss $\mathcal{L}_\mathrm{L}$ and $\mathcal{L}_\mathrm{AN}$ under the similar numerical value with classification losses $\mathcal{L}_\mathrm{cls}^\mathrm{S}$ and $\mathcal{L}_\mathrm{cls}^\mathrm{S}$. Moreover, all the feature channels are set to be $512$ for $\mathbf{h}_{t}^{l}$, $\mathbf{a}_{t}^{l}$, and $\mathbf{g}_{t}^{l}$. 

\noindent \textbf{Features.} To make comparisons with the previous works that process the video with extracted features, we also extract the two-stream features for these two datasets. The dimension for both RGB and optical flow is 1024. The two-stream input features $\mathbf{x}_{t}$ concatenate the RGB and optical flow features. The toolkit for extracting features is provided by \cite{liu2019completeness} and the extractor is pre-trained on Kinetics~\cite{carreira2017quo}.

\begin{table}[t]
\centering
\caption{Comparisons to state-of-the-art methods based on two datasets: THUMOS14 and TVSeries. $\ast$ means the feature extractor is pre-trained on Kinetics dataset.}
\label{table:sota}
\setlength{\tabcolsep}{15pt}{
\begin{tabular}{lcc}
\toprule
\multirow{2}{*}{Method}              & THUMOS14                     & TVSeries        \\ \cline{2-3} 
                                     & \multicolumn{1}{c}{mAP (\%)} & mcAP (\%)       \\ \hline
RED~\cite{gao2017red}                & 45.3                         & 79.2            \\
$\mathrm{F^{2}G}$~\cite{yoon2020a}   & 45.8                         & -               \\
TRN~\cite{xu2019trn}                 & 47.2                         & 83.7            \\
IDN~\cite{eun2020learning}           & 50.0                         & 84.7            \\ \hline
TRN$\ast$~\cite{gao2020woad}         & 51.0                         & -               \\
IDN$\ast$~\cite{eun2020learning}     & 60.3                         & 86.1            \\
Ours$\ast$                           & \textbf{64.5}                & \textbf{86.4}   \\ \bottomrule
\end{tabular}}
\end{table}

\subsection{Comparison to the State-of-the-arts}

In this section, we make comparisons between our proposed method and other state-of-the-art methods on TVSeries and THUMOS14 datasets. As illustrated in Table~\ref{table:sota}, the inputs of $\mathrm{F^{2}G}$~\cite{yoon2020a} are video frames while other methods take extracted two-stream features as the input data.  $\mathrm{F^{2}G}$ is designed to first predict future video frames and then use these predicted frames as well as previous and current video frames to classify current actions. RED~\cite{gao2017red} and TRN~\cite{xu2019trn} also use the anticipation-classification framework for OAD but the prediction stages are based on the feature level which is built in a reinforced encoder-decoder network or an LSTM structure, respectively. IDN~\cite{eun2020learning} is an improved LSTM structure for the OAD task that decides whether to accumulate input information based on its relevance to the current input frame. Besides, $\ast$TRN~\cite{xu2019trn} and $\ast$IDN~\cite{eun2020learning} also use the feature extractor that is pre-trained on Kinetics dataset~\cite{carreira2017quo}, which allows a fair comparison with our proposed method. On the THUMOS14 dataset, our proposed PKD method achieves $64.5\%$ in terms of mAP that improves absolute $4.2\%$ over the previous best method IDN. As for the TVSeries dataset, PKD also performs better than the previous method. Since the performance is relatively high on this dataset, the improvement is only absolute $0.3\%$ in terms of mcAP over the previous best method.

\subsection{Ablation Studies on PKD}

In this section, we first show the effectiveness of introducing KD to the OAD task, then investigate the effectiveness of our proposed curriculum learning process and auxiliary nodes on two datasets: TVSeries and THUMOS14. Besides, we also explore the performance of PKD with different numbers of auxiliary nodes.

\begin{table}[t]
\centering
\caption{Comparisons in terms of mcAP (\%) on the TVSeries dataset. The first two rows show the baseline performance on the online and different offline models. Rest of the table illustrate the effectiveness of different KD paths and KD losses for online $\mathtt{S}$.}
\label{table:TVSeries}
\setlength{\tabcolsep}{7.85pt}{
\begin{tabular}{cccccc}
\toprule
models  & $\mathtt{S}$    & $\mathtt{T}_{1}$    & $\mathtt{T}_{2}$    & $\mathtt{T}_{3}$    & $\mathtt{T}_{4}$    \\ \hline
mcAP    & 85.40 & 86.66 & 87.56 & 87.77 & 87.94 \\ \hline
\end{tabular}}
\setlength{\tabcolsep}{6.7pt}{
\begin{tabular}{c|cc}
\hline
KD path                & $\mathtt{S}$ w/ $\mathcal{L}_\mathrm{L}$ & $\mathtt{S}$ w/ $\mathcal{L}_\mathrm{L}$+$\mathcal{L}_\mathrm{AN}$  \\ \hline
$\mathtt{T}_{1}\to\mathtt{S}$                                                      & 85.57      & 86.07             \\
$\mathtt{T}_{2}\to\mathtt{S}$                                                      & 85.74      & 86.13             \\
$\mathtt{T}_{3}\to\mathtt{S}$                                                      & 85.94      & 86.14             \\
$\mathtt{T}_{4}\to\mathtt{S}$                                                      & 85.98      & 86.14             \\
$\mathtt{T}_{4}\to\mathtt{T}_{2}\to\mathtt{S}$                                     & 85.73      & 86.20             \\
$\mathtt{T}_{4}\to\mathtt{T}_{3}\to\mathtt{T}_{2}\to\mathtt{T}_{1}\to\mathtt{S}$   & 85.96      & 86.30             \\
$\mathtt{S}\to\mathtt{T}_{2}\to\mathtt{T}_{4}$                                     & 86.11      & 86.32             \\
$\mathtt{S}\to\mathtt{T}_{1}\to\mathtt{T}_{2}\to\mathtt{T}_{3}\to\mathtt{T}_{4}$   & 86.43      & \textbf{86.44}    \\ \bottomrule
\end{tabular}}
\end{table}

\noindent \textbf{Effectiveness of the KD.}
As illustrated in Table~\ref{table:TVSeries} and Table~\ref{table:THUMOS14}, the first two rows of each table show the performances on $\mathtt{S}$, $\mathtt{T}_{1}$, $\mathtt{T}_{2}$, $\mathtt{T}_{3}$, and $\mathtt{T}_{4}$ models of each dataset. The baseline online model $\mathtt{S}$ achieves $85.40\%$ and $61.65\%$ in terms of the mcAP on TVSeries and the mAP on THUMOS14, respectively. While the best offline model $\mathtt{T}_{4}$ can achieve $87.94\%$ and $66.91\%$. So the performance gaps between the online and the offline models are absolute $2.54\%$ on TVSeries and $5.26\%$ on THUMOS14, which shows the upper bounds for online models through distilling knowledge from offline models. We denote ``$\mathtt{T}_{1}\to\mathtt{S}$'', ``$\mathtt{T}_{2}\to\mathtt{S}$'', ``$\mathtt{T}_{3}\to\mathtt{S}$'', and ``$\mathtt{T}_{4}\to\mathtt{S}$'' as the one step distillation settings which means the online model $\mathtt{S}$ is trained from only one offline model.  The results show that the student model $\mathtt{S}$ obtains a better performance when it learns from a better teacher who can see more future frames. Therefore, the baseline KD for OAD that only uses the vanilla KD loss $\mathcal{L}_\mathrm{L}$ in ``$\mathtt{T}_{4}\to\mathtt{S}$'' improves the performance of $\mathtt{S}$ from $85.40\%$ to $85.98\%$ on TVSeries dataset and from $61.65\%$ to $62.72\%$ on THUMOS14 dataset. Considering the upper bounds from offline models. The effectiveness of the baseline KD for OAD can reduce this gap by $22.8\%$ on TVSeries and $20.3\%$ on THUMOS14.

\begin{table}[t]
\centering
\caption{Comparisons in terms of mAP (\%) on the THUMOS14 dataset. The first two rows show the baseline performance on the online and different offline models. Rest of the table illustrate the effectiveness of different KD paths and KD losses for model $\mathtt{S}$.}
\label{table:THUMOS14}
\setlength{\tabcolsep}{7.85pt}{
\begin{tabular}{cccccc}
\toprule
models  & $\mathtt{S}$    & $\mathtt{T}_{1}$    & $\mathtt{T}_{2}$    & $\mathtt{T}_{3}$    & $\mathtt{T}_{4}$    \\ \hline
mAP     & 61.65 & 63.92 & 64.91 & 65.75 & 66.91 \\ \hline
\end{tabular}}
\setlength{\tabcolsep}{6.7pt}{
\begin{tabular}{c|cc}
\hline
KD path                          & $\mathtt{S}$ w/ $\mathcal{L}_\mathrm{L}$ & $\mathtt{S}$ w/ $\mathcal{L}_\mathrm{L}$+$\mathcal{L}_\mathrm{AN}$ \\ \hline
$\mathtt{T}_{1}\to\mathtt{S}$                                                      & 62.02      & 62.94         \\
$\mathtt{T}_{2}\to\mathtt{S}$                                                      & 62.35      & 63.23         \\
$\mathtt{T}_{3}\to\mathtt{S}$                                                      & 62.70      & 63.77         \\
$\mathtt{T}_{4}\to\mathtt{S}$                                                      & 62.72      & 63.96         \\
$\mathtt{T}_{4}\to\mathtt{T}_{2}\to\mathtt{S}$                                     & 62.75      & 63.43         \\
$\mathtt{T}_{4}\to\mathtt{T}_{3}\to\mathtt{T}_{2}\to\mathtt{T}_{1}\to\mathtt{S}$   & 62.35      & 63.47         \\
$\mathtt{S}\to\mathtt{T}_{2}\to\mathtt{T}_{4}$                                     & 62.89      & 64.31         \\
$\mathtt{S}\to\mathtt{T}_{1}\to\mathtt{T}_{2}\to\mathtt{T}_{3}\to\mathtt{T}_{4}$   & 63.64      & \textbf{64.45}             \\ \bottomrule
\end{tabular}}
\end{table}

\noindent \textbf{Effectiveness of the auxiliary nodes.} Apart from distilling the knowledge from the output logits, we also design auxiliary nodes for $\mathtt{S}$ to distill the privileged knowledge through feature representations. As shown in Table~\ref{table:TVSeries} and Table~\ref{table:THUMOS14}, with adding the loss $\mathcal{L}_\mathrm{AN}$, the performance on both datasets is improved on different KD path. In ``$\mathtt{T}_{4}\to\mathtt{S}$'' setting, the performance can achieve $86.14\%$ on TVSeries and $63.96\%$ on THUMOS14. Therefore, the effectiveness of the auxiliary nodes can further reduce the upper bound gap to $29.1\%$ on TVSeries and $43.9\%$ on THUMOS14.

\noindent \textbf{Effectiveness of the curriculum learning.} As for KD with more distillation path, the notations ``$\mathtt{T}_{4}\to\mathtt{T}_{2}\to\mathtt{S}$'' and ``$\mathtt{T}_{4}\to\mathtt{T}_{3}\to\mathtt{T}_{2}\to\mathtt{T}_{1}\to\mathtt{S}$'' follow the TAKD \cite{mirzadeh2020improved} training strategy that train the online model $\mathtt{S}$ starting from a strong teacher model $\mathtt{T}_{4}$; then distill the knowledge to some intermediate teacher models; finally, the student model $\mathtt{S}$ is taught with the closest teacher model. As for the notations ``$\mathtt{S}\to\mathtt{T}_{2}\to\mathtt{T}_{4}$'' and ``$\mathtt{S}\to\mathtt{T}_{1}\to\mathtt{T}_{2}\to\mathtt{T}_{3}\to\mathtt{T}_{4}$'' represent the curriculum learning process for PKD in Algorithm~\ref{alg:bwpkd}. 
As shown in Table~\ref{table:TVSeries} and Table~\ref{table:THUMOS14}, ``$\mathtt{T}_{4}\to\mathtt{T}_{2}\to\mathtt{S}$'' and ``$\mathtt{T}_{4}\to\mathtt{T}_{3}\to\mathtt{T}_{2}\to\mathtt{T}_{1}\to\mathtt{S}$'' on both datasets can improve the performance from the baseline model $\mathtt{S}$. However, they obtain about the same or even worse results as the single path ``$\mathtt{T}_{4}\to\mathtt{S}$'' dose. TAKD has been proved to be effective in conventional KD settings where the knowledge gap mainly comes from the model structure. So it can alleviate the ability of feature representations from strong to weak. But in offline to online settings, the representation ability of student $\mathtt{S}$ is not weak compared to teacher $\mathtt{T}_{4}$. So, KD in TAKD way loses the input information since the final knowledge distillation stage is completed by a weaker teacher. Therefore, it is proper to let the student approach the teacher instead of letting the teacher approach the student. Following the training strategy in Algorithm~\ref{alg:bwpkd}, the curriculum learning procedure ``$\mathtt{S}\to\mathtt{T}_{1}\to\mathtt{T}_{2}\to\mathtt{T}_{3}\to\mathtt{T}_{4}$'' achieves the best performance on both datasets, which can improve the online model $\mathtt{S}$ from $85.40\%$ to $86.44\%$ on TVSeries and from $61.65\%$ to $64.45\%$ on THUMOS14. Therefore, the effectiveness of the curriculum learning can further reduce the upper bound gap to $40.9\%$ on TVSeries and $53.2\%$ on THUMOS14.

\begin{figure}[t]
\begin{center}
  \includegraphics[width=0.9\linewidth]{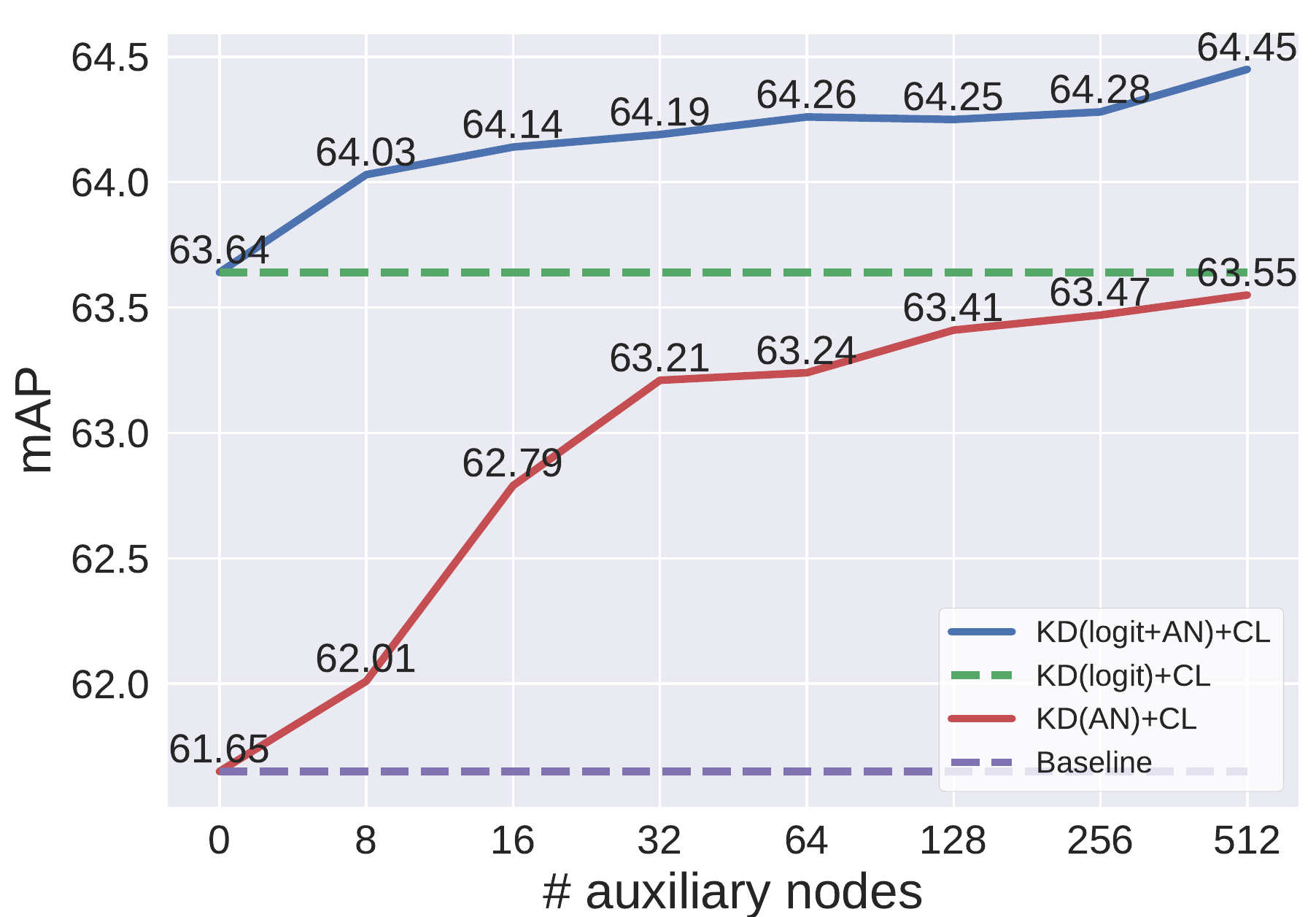}
\end{center}
\caption{Comparisons between different number of auxiliary nodes in terms of mAP on the THUMOS14 dataset.
}
\label{fig:an}
\end{figure}

\noindent \textbf{The number of auxiliary nodes.}  As illustrated in Figure~\ref{fig:an}, we also investigate the performance of PKD with different numbers of auxiliary nodes based on ``$\mathtt{S}\to\mathtt{T}_{1}\to\mathtt{T}_{2}\to\mathtt{T}_{3}\to\mathtt{T}_{4}$'' setting. In previous experiments, we introduce auxiliary nodes to each feature layer of $\mathtt{S}$ which has the same feature dimension as the teacher. To match the feature dimension with the teacher when changing the number of auxiliary nodes, we use an additional causal convolution layer to map this dimension difference. Results show that PKD achieves better performance with more auxiliary nodes. However, only using $16$ auxiliary nodes can also improve $1.14\%$ mAP from the ``Baseline'' model.

\subsection{Predicting Features vs. PKD}

To compare our proposed PKD with anticipation methods, we also implement an anticipation\&classification pipeline with the same network structure $\mathtt{S}$. The anticipation stage is made under the input feature level $\mathbf{x}_{t}$. It is designed like the proposed auxiliary nodes that we predict input features supervised by different average future inputs. For each input feature $\mathbf{x}_{t}$, the predicted input feature is notated as $\hat{\mathbf{x}}_{t}$. Thus, $\hat{\mathbf{x}}_{t}$ and $\mathbf{x}_{t}$ are concatenated as the new input features for time $t$.  $\hat{\mathbf{x}}_{t}$ is supervised as:

\begin{equation}\label{eq:predict}
    \mathcal{L}_{\mathrm{predict}}=||\hat{\mathbf{x}}_{t}-\frac{1}{P}\sum_{i=1}^{P}\mathbf{x}_{t+i}||^{2}.
\end{equation}

\begin{figure}[t]
\begin{center}
  \includegraphics[width=0.9\linewidth]{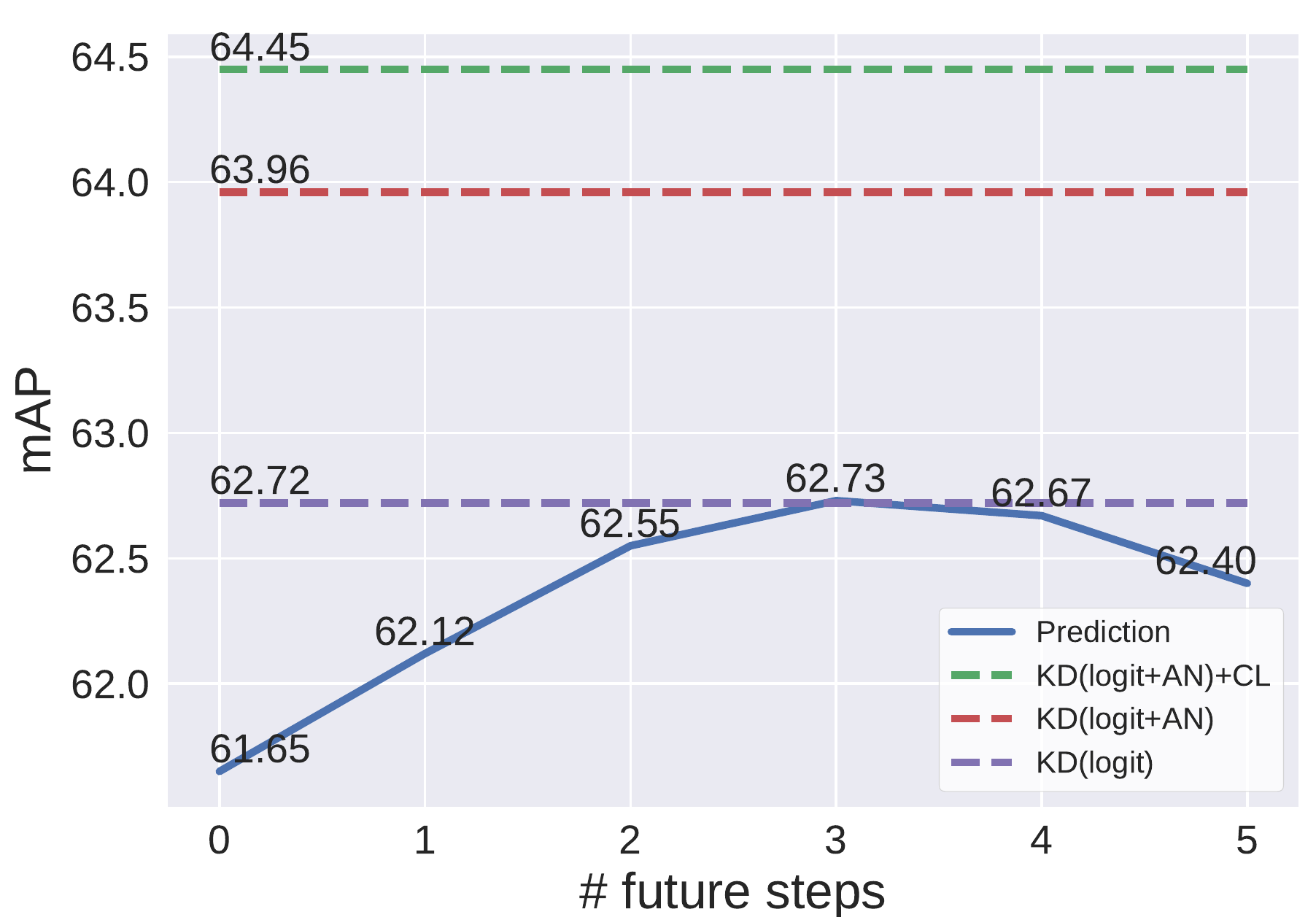}
\end{center}
\caption{Comparisons between prediction method and KD methods in terms of mAP on the THUMOS14 dataset.
}
\label{fig:predict}
\end{figure}

As illustrated in Figure~\ref{fig:predict}, the performance on the THUMOS14 dataset of the prediction method is shown in a blue solid line with different average future steps $P$, while the KD methods are shown in dashed lines. By predicting more future features the online model can achieve better performance but when the average future step is large than $3$, the performance decreases. The phenomenon is the same as the prediction method TRN~\cite{xu2019trn} which also shows that a larger number of prediction steps do not guarantee better performance. It is because anticipation accuracy usually decreases for longer future sequences, and thus creates more noise in the input features. At the average step of $3$, the anticipation method achieves the best performance of $62.73\%$. Comparing the prediction method and KD methods, we find that the prediction method can achieve a similar performance of baseline KD. However, this explicit anticipation way still has an obvious gap around the absolute $1.7\%$ between our proposed PKD.

\begin{table*}[t]
\centering
\small
\caption{Comparisons on different portions of the action instance in terms of mcAP on the TVSeries dataset. The portion labels are only used in evaluation when the results is predicted by an OAD model. $\ast$ means the feature extractor is pre-trained on Kinetics.}
\label{table:TVSeries_proportion}
\setlength{\tabcolsep}{3.8pt}{
\begin{tabular}{ccccccccccc}
\toprule
\multirow{2}{*}{Method}             & \multicolumn{10}{c}{Proportion of Action}                                                                              \\ \cline{2-11} 
                                    & 0\%-10\% & 10\%-20\% & 20\%-30\% & 30\%-40\% & 40\%-50\% & 50\%-60\% & 60\%-70\% & 70\%-80\% & 80\%-90\% & 90\%-100\% \\ \hline
CNN~\cite{geest2016online}          & 61.0     & 61.0      & 61.2      & 61.1      & 61.2      & 61.2      & 61.3      & 61.5      & 61.4      & 61.5       \\
LSTM~\cite{geest2016online}         & 63.3     & 64.5      & 64.5      & 64.3      & 65.0      & 64.7      & 64.4      & 64.4      & 64.4      & 64.3       \\
FV-SVM~\cite{geest2016online}       & 67.0     & 68.4      & 69.9      & 71.3      & 73.0      & 74.0      & 75.0      & 75.4      & 76.5      & 76.8       \\
TRN~\cite{xu2019trn}                & 78.8     & 79.6      & 80.4      & 81.0      & 81.6      & 81.9      & 82.3      & 82.7      & 82.9      & 83.3       \\
IDN~\cite{eun2020learning}          & 80.6     & 81.1      & 81.9      & 82.3      & 82.6      & 82.8      & 82.6      & 82.9      & 83.0      & 83.9       \\ \hline
IDN$\ast$~\cite{eun2020learning}    & 81.7     & 81.9      & 83.1      & 82.9      & 83.2      & 83.2      & 83.2      & 83.0      & 83.3      & 86.6       \\ 
Ours$\ast$                          & \textbf{82.1}     & \textbf{83.5}      & \textbf{86.1}      & \textbf{87.2}      & \textbf{88.3}      & \textbf{88.4}      & \textbf{89.0}      & \textbf{88.7}      & \textbf{88.9}      & \textbf{87.7}       \\ \bottomrule
\end{tabular}}
\end{table*}

\begin{figure*}[t]
\begin{center}
  \includegraphics[width=1.0\linewidth]{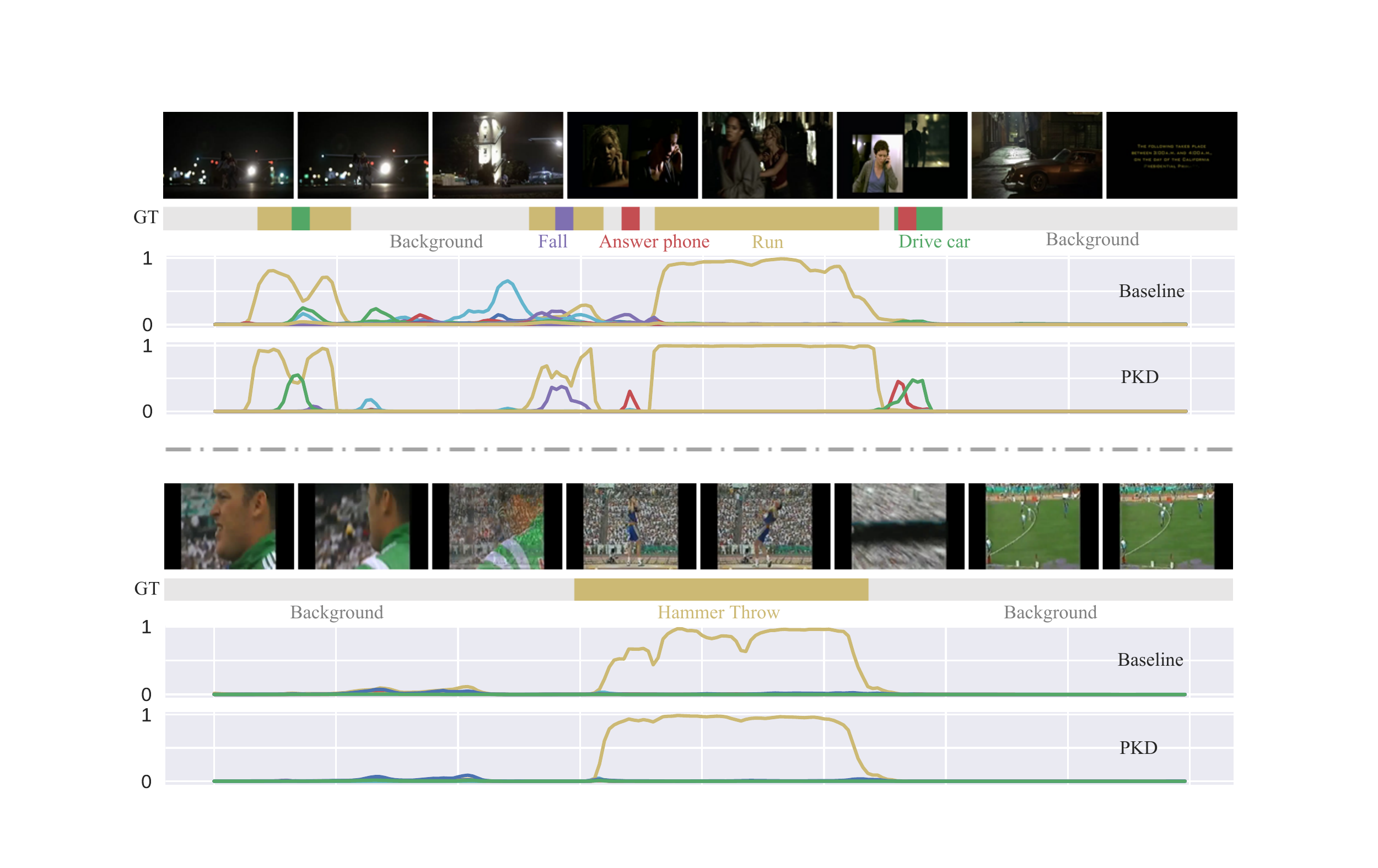}
  
\end{center}
\caption{Comparisons between the PKD and the baseline OAD model $\mathtt{S}$ in terms of mAP on TVSeries (above) and THUMOS14 (below).
}
\label{fig:visualization}
\end{figure*}

\subsection{Evaluation of Different Stages of Action}

One of the significant characteristics of OAD is to detect an action at the early stage. Therefore, we compare our proposed PKD method with the previous methods in different portions of action stages on the TVSeries dataset in Table~\ref{table:TVSeries_proportion}. For example, ``$0\%$-$10\%$'' means that only action frames at the duration range of ``$0\%$-$10\%$'' are evaluated in terms of mcAP. The results show that our proposed PKD outperforms all the previous methods at all action portions. We notice that CNN~\cite{geest2016online} and LSTM~\cite{geest2016online} obtain almost the same result on different action portions since they do not utilize the temporal information. \emph{i.e.}, these methods only use a single frame or few successive frames at each output time. As for FV-SVM~\cite{geest2016online}, TRN~\cite{xu2019trn}, and IDN~\cite{eun2020learning} which use the accumulated information, the performance increases when observing more completed actions. Different from previous methods, our PKD takes the 1D temporal convolution as our base model that processes the input sequence within a receptive window, thus reaches its maximum $89.0\%$ near the middle and back of the action.

\subsection{Qualitative Results}

As illustrated in Figure~\ref{fig:visualization}, we visualize some OAD results on TVSeries (above) and THUMOS14 (below) datasets. Comparing the baseline model $\mathtt{S}$ with our proposed PKD method, we find that PKD can detect more action instances that have been missed in the baseline model the on TVSeries dataset. \emph{e.g.}, the actions ``Fall'', ``Answer phone'', and ``Drive car'' are missed by the baseline model but found by the PKD method. It is because that the PKD model obtains the privileged knowledge from the offline models, which can help to classify the ongoing actions. Besides, the PKD method can output more stable predictions than the baseline which are shown inside the ``Hammer Throw'' action.


\section{Conclusions}

In this paper, we propose a novel algorithm that applies knowledge distillation (KD) to online action detection (OAD) so that the target model is guided by a teacher model that sees complete video data. The key difficulty lies in that teacher and student models receives different input data. To bridge the information gap, we present a privileged KD pipeline that leverages curriculum learning and inserts auxiliary nodes to assist the student model. Experiments on TVSeries and THUMOS14 demonstrate the state-of-the-art performance of our approach.

Our research enlightens a new path that improves video understanding by offering auxiliary guidance (\textit{e.g.}, predicting unseen features) in an implicit space. This is probably related to unsupervised of self-supervised representation learning, which we will investigate in the future.

{\small
\bibliographystyle{ieee_fullname}
\bibliography{egbib}
}

\end{document}